\documentclass[10pt,twocolumn,letterpaper]{article}

\usepackage{graphicx}
\usepackage{amsmath}
\usepackage{amssymb}
\usepackage{booktabs}

\usepackage[pagebackref,breaklinks,colorlinks]{hyperref}

\usepackage[capitalize]{cleveref}
\crefname{section}{Sec.}{Secs.}
\Crefname{section}{Section}{Sections}
\Crefname{table}{Table}{Tables}
\crefname{table}{Tab.}{Tabs.}

\def\cvprPaperID{*****} 
\def\confName{CVPR}
\def\confYear{2023}

\begin{document}

\title{Enhanced Self-Checkout System for Retail Based on Improved YOLOv10}



\author{
  \begin{tabular}{ c c }
       Lianghao Tan                     &    Shubing Liu                                 \\        
       Arizona State University         &    University of North Carolina at Chapel Hill \\ 
       \\
       Jing Gao                         &    Xiaoyi Liu                                  \\
       University of Minnesota          &    Arizona State University                    \\
       \\ 
       Linyue Chu                       &    Huangqi Jiang                               \\
       University of California, Irvine &    Case Western Reserve University
  \end{tabular}
}

\maketitle


\raggedbottom
\begin{abstract}
With the rapid advancement of deep learning technologies, computer vision has shown immense potential in retail automation. This paper presents a novel self-checkout system for retail based on an improved YOLOv10 network, aimed at enhancing checkout efficiency and reducing labor costs. We propose targeted optimizations to the YOLOv10 model, by incorporating the detection head structure from YOLOv8, which significantly improves product recognition accuracy. Additionally, we develop a post-processing algorithm tailored for self-checkout scenarios, to further enhance the application of system. Experimental results demonstrate that our system outperforms existing methods in both product recognition accuracy and checkout speed. This research not only provides a new technical solution for retail automation but  offers valuable insights into optimizing deep learning models for real-world applications. 
\end{abstract}

\section{Introduction}
\label{sec:intro}
\subsection{Research Background}

Efficient retail operations are critical to ensuring customer satisfaction and maintaining competitive advantage in the fast-paced market today. Self-checkout systems, in particular, have become an essential component of modern retail environments, offering customers a quicker, more convenient way to complete their purchases \cite{orel2014supermarket}. These systems help reduce the burden on staff, minimize wait times, and enhance the overall shopping experience. As consumer expectations for speed and convenience rise, the demand for reliable and accurate self-checkout solutions continues to grow. However, the effectiveness of these systems hinges on their ability to accurately identify and process a diverse range of products in real-time, a challenge that has driven significant research and development in the field of computer vision and deep learning.


\subsection{Challenges in Current Self-Checkout Systems}

Despite the growing adoption of self-checkout systems in modern retail environments, several challenges persist that can hinder their effectiveness. One of the primary issues is the accurate recognition and differentiation of a wide variety of products, especially in environments with complex backgrounds, varying lighting conditions, or occluded items \cite{vats2023enhancing}. Traditional systems often struggle with identifying items that are similar in appearance or those that are irregularly shaped, leading to increased error rates and customer frustration. Additionally, maintaining real-time processing capabilities without compromising accuracy remains a significant hurdle, particularly as the volume of transactions and the diversity of products continue to expand \cite{shoman2022region}. These challenges underscore the need for ongoing research and development in computer vision and deep learning to create more robust and reliable self-checkout systems that can meet the demands of today’s retail industry.

\subsection{Overview of Advancements of Deep Learning and Computer Vision in Retail}


With the rapid advancement of technology, the wave of Artificial Intelligence (AI) is profoundly transforming a wide range of industries, including retail\cite{oosthuizen2021artificial,zheng2024advanced}. Both 2-D and 3-D vision models are widely used in different scenarios\cite{liu2024digital,lin2024neural} and real-time tracking detection can be used on both manned and unmanned objects\cite{liu2023siamman,mokayed2023real}. In the realm of AI, Machine Learning (ML) has become a powerful tool for tackling complex tasks and enhancing efficiency. Deep Learning (DL) and Computer Vision (CV), two significant subsets of ML, have gained attention in recent years for their precise handling and analysis of visual data even in difficult and remote scenarios\cite{he2023camouflaged,li2024lr}. 

The development of powerful Convolutional Neural Networks (CNNs) and real-time object detection algorithms, such as the YOLO (You Only Look Once) series, has enabled machines to process visual data with unprecedented speed and accuracy \cite{gu2018recent,redmon2016you}. These technologies have greatly improved the ability to recognize and classify retail products in various environments, even under challenging conditions such as occlusions, varying lighting, and complex backgrounds.


\subsection{Purpose of the New System and Advantages}

The purpose of the current study is to develop an enhanced self-checkout system that effectively addresses the limitations of existing technologies by leveraging the latest advancements in deep learning and computer vision. Our proposed system, MidState-YOLO-ED, integrates key innovations from YOLOv10 with enhancements inspired by YOLOv8, aiming to significantly improve product recognition accuracy and processing efficiency in real-time retail environments \cite{wang2023uav, wang2024yolov10}. The advantages of this system include its ability to accurately identify a diverse range of retail products, maintain robust performance under challenging conditions, and operate efficiently on resource-constrained devices. These improvements are designed to make the self-checkout process faster, more reliable, and better suited to the demands of modern retail settings.

\section{Related Work}

\subsection{Traditional Object Detection Algorithms}

Object detection has long been a central challenge in computer vision, with traditional approaches primarily relying on CNNs to identify and locate objects within images. Among the most notable early models are Faster R-CNN, Single Shot MultiBox Detector (SSD), and the original YOLO series. Before the advent of DL, the field of object detection primarily relied on traditional manual feature engineering techniques while DL advanced the object detection and brought the benefit to retails and consumers \cite{wei2020deep,dang2024real}. Solutions from this era typically followed a two-stage process: feature extraction and classification. Besides the advancement of algorithms for exploring data samples \cite{tarvainen2017mean,duan2022mutexmatch}, techniques from image processing and computer vision were used to manually construct and extract prominent features from images, such as Haar feature sets and Scale-Invariant Feature Transform (SIFT)\cite{cruz2012scale}. These designed features were then fed into classifiers to identify objects within images. However, the effectiveness of this approach heavily depended on the quality of the feature engineering and often fell short in complex and variable image backgrounds and scenes.  

With the widespread adoption of CNNs, a significant leap occurred in object detection. CNNs, with their robust automatic feature-learning capabilities, eliminated the need for cumbersome manual feature design\cite{tokunaga2019adaptive,jin2024graphcnnpred}. They could directly learn hierarchical and rich feature representations from raw image data, greatly enhancing the accuracy and efficiency of object detection. Faster R-CNN marked a significant milestone by introducing the Region Proposal Network (RPN), which efficiently generates region proposals that are likely to contain objects, reducing the need for exhaustive search methods and improving detection accuracy \cite{girshick2015fast}. However, its two-stage process, which separates region proposal and classification, makes it computationally intensive, limiting its real-time application.

SSD aimed to balance accuracy and speed by eliminating the region proposal stage entirely. Instead, it predicts object classes and bounding boxes directly from feature maps at different scales, allowing for the detection of objects of various sizes in a single pass \cite{liu2016ssd}. While SSD offers improved speed over Faster R-CNN, it still struggles with detecting smaller objects and achieving the highest levels of accuracy \cite{li2017fssd}.

YOLO introduced a groundbreaking approach by framing object detection as a single regression problem, predicting bounding boxes and class probabilities directly from the entire image in one forward pass through the network. This made YOLO exceptionally fast compared to its predecessors, making it suitable for real-time applications. However, early versions of YOLO had limitations in detecting small and overlapping objects and struggled with localization accuracy \cite{jiang2022review}.

These traditional algorithms laid the groundwork for the rapid advancements in object detection seen in later models. They highlighted the trade-offs between speed and accuracy and set the stage for more sophisticated approaches, such as the newer iterations of the YOLO series, which aim to overcome these limitations.

\subsection{Development of the YOLO Series}

The YOLO series has significantly influenced the field of object detection since its introduction \cite{redmon2016you}. By reimagining object detection as a single regression problem, YOLO enables real-time processing with relatively high accuracy, making it a groundbreaking approach in computer vision.

YOLOv1 laid the foundation by dividing an input image into a grid and predicting bounding boxes and class probabilities directly from each grid cell \cite{redmon2016you}. While this approach offered unprecedented speed, it struggled with detecting small and overlapping objects due to the limitations of its grid-based prediction method. In response to these challenges, YOLOv2 introduced anchor boxes, which improved the accuracy of bounding box predictions by allowing the model to predict multiple bounding boxes for each grid cell \cite{sang2018improved}. YOLOv2 also adopted the Darknet-19 backbone, significantly enhancing its feature extraction capabilities. The introduction of multi-scale training further allowed the model to generalize better across different object sizes and shapes, making YOLOv2 a more robust solution.

Furthermore, YOLOv3 incorporated a multi-scale feature pyramid network (FPN), which enabled better detection of objects at varying scales by merging features from different layers of the network \cite{redmon2018yolov3}. YOLOv3 also improved class prediction accuracy, enhancing the model's ability to recognize objects across a broader range of categories. YOLOv4 and YOLOv5 continued to refine the architecture with a focus on optimizing both speed and accuracy \cite{bochkovskiy2020yolov4, zhu2021tph}. These versions integrated better backbone networks, more advanced feature fusion techniques, and improved loss functions. The enhancements made in these iterations further solidified YOLO's position as a leading framework for real-time object detection, particularly in scenarios requiring a balance between computational efficiency and detection performance.

With the introduction of YOLOv8, the series saw significant architectural innovations, including the use of Cross-Stage Partial Network (CSPNet) for more efficient feature extraction and anchor-free detection heads that simplified the model's design \cite{terven2023comprehensive}. YOLOv8 also leveraged the SiLU activation function, which facilitated better gradient flow during training, leading to faster convergence and higher accuracy.

The most recent version, YOLOv10, represents a culmination of these advancements, introducing a non-maximum suppression (NMS)-free training strategy that minimizes inference delays \cite{wang2024yolov10}. YOLOv10 also employs a dual-label assignment mechanism that enhances both recall and precision, making it the most advanced and capable iteration in the YOLO series.

The continuous evolution of the YOLO series highlights the ongoing efforts to balance speed, accuracy, and efficiency, ensuring that each new version builds on the strengths of its predecessors while addressing their limitations.

\subsection{Application of YOLO}

The YOLO series has seen widespread adoption across various industries due to its real-time object detection capabilities. In autonomous driving, YOLO is used for detecting pedestrians, vehicles, and other road elements, ensuring timely decisions for safety \cite{sarda2021object}. In surveillance and security, it aids in monitoring environments, detecting intruders, and analyzing crowd behavior, making it ideal for real-time threat detection \cite{narejo2021weapon}. 

YOLO’s applications extend to healthcare, where it assists in detecting abnormalities in medical images \cite{qureshi2023comprehensive}. In agriculture, YOLO is employed in precision farming to monitor crop health and detect pests, helping optimize yields \cite{li2020agricultural}. In public service, YOLO is used to enhance automatic pavement distress recognition to assist highway maintenance decision-making \cite{dan2024multiple}. In the retail industry, YOLO powers automated checkout systems by accurately identifying products, enhancing customer experience, and streamlining inventory management \cite{vats2023enhancing}. The versatility and efficiency of YOLO across these diverse applications highlight its significant impact on real-time object detection across multiple sectors.

\section{The Improved MidState-YOLO-ED Network}
\subsection{Integration of YOLOv8 and YOLOv10}
One of the most distinct features of YOLOv10 compared to its predecessors is the elimination of Non-Maximum Suppression (NMS), achieved by introducing a consistent dual-assignment strategy \cite{wang2024yolov10}. This strategy involves calling the loss function calculation method of YOLOv8 twice, summing the results, and returning them. This approach aims to address the issue of redundant predictions in post-processing, aligning closer to the end-to-end direction of the RT-DETR model. Unfortunately, this modification led to a decrease in accuracy for many datasets in practical applications. To avoid loss of precision, the prediction head of YOLOv10 was reverted to that of YOLOv8.

To significantly reduce computational redundancy and achieve a more efficient architecture, YOLOv10 employs a lightweight classification head, Spatial-Channel Decoupled Downsampling (SCDD), and a rank-based block design (i.e., C2fUIB). SCDD first adjusts the channel dimensions using point-wise convolution, followed by spatial downsampling using depth-wise convolution to reduce the number of parameters. However, some information loss occurs during the downsampling process in experiments, which, although reducing latency, does not ensure performance gains.

Due to the modifications in this study reverting two core components of YOLOv10 back to YOLOv8 status, and integrating both versions, the current model has been named MidState-YOLO. Additionally, the model cleverly integrates Efficient Multi-Scale Attention (EMA) attention and the C2f-Dual convolution design, ultimately naming the final model as the MidState-YOLO-ED network.

\subsection{Integration of EMA Attention}
To further enhance the expressive capability of the MidState-YOLO network and establish long and short dependency relationships, EMA has been integrated into the Neck network. EMA is a parallel attention mechanism used in computer vision tasks to improve model performance and processing speed. Unlike traditional CNNs, EMA adopts a parallel structure to handle input data. This parallel convolution allows faster training of models when dealing with large-scale data and enhances accuracy by enabling parallel processing of features at different scales. In figure x, the divided groups are represented by "g," while "X Avg Pool" denotes one-dimensional horizontal global pooling, and "Y Avg Pool" represents one-dimensional vertical global pooling respectively.
The formula for the average pooling operation is as follows, where \textit{Xc(i, j)} represents the element at position \textit{(i, j)}:
\begin{equation}
Z_c = \frac{1}{H \times W}\sum_{j}^{H}\sum_{i}^{W}X_c\left ( i, j \right )
  \label{eq:important}
\end{equation}   

The input to EMA is first grouped and reshaped, redistributing part of the channel dimensions to the batch dimension. This is followed by further subdivision of the channel dimension into multiple sub-features to preserve key information in each channel and optimize the distribution of spatial semantic features\cite{zhong2024fagd}.

\begin{figure*}
\centering
\includegraphics[width=\textwidth]{"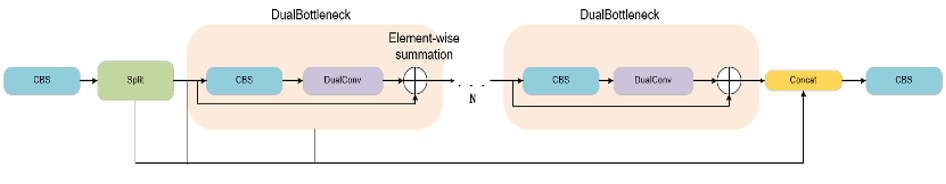"}
\caption{Dual Convolution C2f-Dual Design}
\end{figure*}

This structure contains two main parallel branches: one branch performs one-dimensional global pooling operations to encode global information, while the other branch performs feature extraction through a 3x3 convolution. The output features from these two branches are modulated through a sigmoid function and normalization processes, and then merged through a cross-dimensional interaction module to capture pixel-level pairwise relationships. Finally, the sigmoid-modulated output feature maps are used to enhance or weaken the original input features, thus achieving a more refined and effective feature representation. Therefore, EMA not only encodes inter-channel information to adjust the importance of various channels but also preserves precise spatial structural details within these channels.

\subsection{Lightweight Dual Convolution C2f-Dual Design}
In YOLOv8 and YOLOv10, the C2f module integrates both low-level and high-level feature maps, facilitating the capture of gradient information flows. However, with the increasing number of layers in CNNs, semantic information in feature maps tends to be progressively extracted and aggregated, leading to redundancy in deep feature maps. Additionally, due to the weight-sharing mechanism of convolutional layers, convolutional kernel parameters are shared at different positions in the deep feature maps, further exacerbating redundancy. Bottleneck modules, composed of many complex convolutions, significantly increase parameter size and computational complexity.

To address this issue, the C2f-Dual convolution design, improved using Dual Convolutional kernels (DualConv), significantly reduces computational costs and the number of parameters while also enhancing precision. This improvement involves replacing the C2f modules before the Spatial Pyramid Pooling (SPPF) with C2f-Dual modules, as shown in Figure 1. This adaptation not only streamlines the network but also optimizes performance by ensuring that critical spatial and semantic features are efficiently processed and integrated.

DualConv is designed to build lightweight deep neural networks by combining 3×3 and 1×1 convolution kernels to process the same input feature map channels, optimizing information processing and feature extraction. In DualConv, the $3\times3$ convolution kernels are used to extract spatial features from the feature maps, capturing more spatial information, while the $1\times1$ convolution kernels integrate these features and reduce the model's parameters. Each group of convolution kernels processes a portion of the input channels independently before the outputs are merged, facilitating efficient flow and integration of information across different feature map channels.

Additionally, DualConv employs group convolution technology to efficiently arrange convolution filters. In group convolution, both input and output feature maps are divided into multiple groups, with each group’s convolution filters processing only a part of the corresponding input feature map. This arrangement allows different kernels within a group to process the same set of input channels in parallel, optimizing information flow and feature extraction efficiency while maintaining the network's representational capabilities.

Thus, replacing the bottleneck structures in C2f with DualBottleneck enriches gradient flow representation, enhances feature extraction capabilities, and reduces the diversity of false positives and false negatives in network learning. This makes it more suitable for retail commodity object detection scenarios.

\section{Experimental Results and Analysis}
\subsection{Experimental Setup and Parameters}
The hardware environment and software configurations used for the experiments are detailed in Table 1. During the model training process, the learning rate was set to 0.01, with optimization carried out using Stochastic Gradient Descent (SGD). The momentum parameter was set at 0.937, and the weight decay factor was 0.0005. The batch size used was 32, and the image size was $640\times640$ pixels. The comparative experiments were conducted over 30 epochs, while the ablation studies were carried out over 25 epochs.

\subsection{Dataset}
This study employs a portion of the retail product (RPC) dataset for training and validation. The RPC dataset, developed by Megvii Technology's Nanjing Research Institute, is currently the largest product recognition dataset available \cite{wei2019rpc}. It includes up to 200 different product categories and totals 83,000 images, realistically simulating retail environments and surpassing existing datasets in fidelity. Moreover, it effectively captures the fine-grained characteristics inherent in the Automatic Check-Out (ACO) problem.

The conceptual approach of this study may differ from that of the researchers who collected the RPC dataset. When customers enter a store and place the items they wish to purchase on the checkout counter, an ideal ACO system would automatically identify each product and accurately generate a shopping list in one go, as shown in Figure 2. Thus, ACO is fundamentally a system designed to identify and count the occurrence of each item in any combination of products. Due to the continuous updating of a vast number of product categories, to avoid exhausting all possible combinations, a feasible solution adopted by the RPC dataset is to collect images of a single product type in a specific environment and reuse them in actual settlements. 

This research posits that there are multiple important metrics for assessing performance on ACO tasks. To ensure accuracy and performance, the images used to train the ACO recognition system should mirror the actual retail checkout environments, which can indeed be simplified and stabilized. Additionally, initial models do not need to exhaust all product combinations to perform ACO tasks; instead, creating random groups of product combinations suffices.

Therefore, for this study, only the checkout configurations from the RPC dataset images are used. We randomly divided 30,000 images of checkout configurations into training, validation, and test sets in an 8:1:1 ratio. This approach aims to provide a realistic yet controlled set of data that reflects real-world ACO system operations while maintaining manageable complexity and variety in training scenarios.

\begin{table}[]
  \caption{Hardware and Software Configuration for Experimental Environment}
  \label{tab:table1}
  \begin{tabular}{lc}
  \hline
    Name & Parameter\\
  \hline
  GPU	&                       RTX4080-16G \\
  CPU	&                       AMD Ryzen7 \\
  Operating System &	        Windows11 \\
  Deep Learning Frameworks &	Pytorch2.1.1\\
                           &   +cuda12.1 \\
  Build System &	            PyCharm \\
  \hline
  \end{tabular}
\end{table}

\begin{figure*}
\centering
\includegraphics[]{"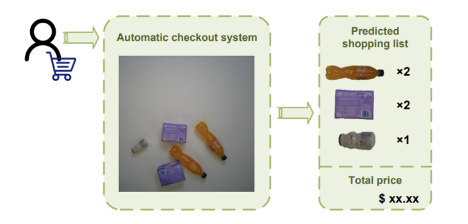"}
\caption{Dual Convolution C2f-Dual Design}
\end{figure*}

\subsection{Evaluation Metrics}
This paper employs Precision (P), Recall (R), mean Average Precision (mAP), number of parameters (Params), and Floating Point Operations (FLOPs) as evaluation metrics, with a set IoU threshold of 0.5. mAP@0.5 denotes the mean Average Precision when the IoU is set at 0.5, and mAP@0.5:0.95 indicates the mean Average Precision when the IoU ranges from 0.5 to 0.95, with a step size of 0.05. The floating point operations indicate the complexity of the algorithm. The specific meanings of other performance metrics are as follows:

\begin{equation}
P = (\frac{TP}{TP + FP}) \times 100%
  \label{eq:important}
\end{equation}   

\begin{equation}
R = (\frac{TP}{TP + FN}) \times 100%
  \label{eq:important}
\end{equation}   

Precision is the probability that a positive sample predicted by the model is indeed a positive sample, and recall is the probability that a positive sample in reality is predicted as positive by the model. These are two important metrics used to evaluate model performance. The expressions are as follows: 

\begin{equation}
AP = \int_{0}^{1} p(r)dr   
  \label{eq:important}
\end{equation}   

\begin{equation}
mAP = \frac{1}{n_j \sum_{j=1}^{n_j}AP_j} 
  \label{eq:important}
\end{equation}   

In the formulas, \textit{TP} (True Positives) refers to positive examples correctly identified as positive by the model; \textit{FP} (False Positives) refers to negative examples incorrectly identified as positive; \textit{FN} (False Negatives) refers to positive examples incorrectly identified as negative.

\textit{mAP} represents the mean of the average precision across all object detection categories. AP is the average of precision values at different recall levels. The curve plotted with Precision (P) as the y-axis and Recall (R) as the x-axis is known as the PR curve. \textit{mAP} is then calculated as the average area under the PR curves for all categories. And \textit{n} represents the number of instances in a given category. \textit{APj} represents the detection precision for category \textit{j}.

\subsection{Ablation Study}
To investigate the extent of improvements from three modification schemes, ablation studies were conducted using YOLOv8-n and YOLOv10-n as baseline networks. These experiments were carried out on the experimental dataset without changing the software and hardware environment, with the only parameter change being the reduction of epochs to 25. As indicated in Table 2, the MidState-YOLO network, which integrates modules from YOLOv8 and YOLOv10, achieved a 23.2\% increase in mAP compared to the original baseline network of YOLOv10-n. This suggests that combining two models can harness the advantages of both while avoiding some of their respective shortcomings, thereby enhancing model performance.

\begin{table*}[h]
  \centering
  \caption{Ablation study experiments with improved strategies}
  \label{tab:table1}
  \begin{tabular}{@{}ccccccc@{}}
  \hline
    Model	               & Precision	    & Recall	     & mAP@0.5	     & mAP@0.5:0.95	  & Params	          & GFLOPs \\
  \hline
    YOLOv8-n               & 0.824          & 0.809	         & 0.877         & 0.691          & 3371024           & 9.8 \\
    YOLO v10-n             & 0.551          & 0.595          & 0.61          & 0.481          & \textbf{28858888} &	9.2 \\
    MidState-YOLO          & 0.794          & 0.775	         & 0.842	     & 0.654          & 3405456           & 9.8 \\
    MidState-YOLO+DualConv & 0.84           & 0.816          & 0.883         & 0.686          & 3251856           & 9.4 \\
    MidState-YOLO+EMA      & 0.843          & 0.813          & 0.884	     & \textbf{0.694} &	3408928           & 9.9 \\
    MidState-YOLO-ED       & \textbf{0.847} & \textbf{0.825} & \textbf{0.89} & 0.691          &	3288096           & 9.6 \\
  \hline
  \end{tabular}
\end{table*}

Using the lightweight dual convolution C2f\_Dual module in the MidState-YOLO network significantly reduced the number of parameters, and all performance metrics showed improvements. The reason for this is the reduction in redundant information in the deep feature maps and a decrease in false positives and false negatives during network learning. The inclusion of the EMA module increased the mAP@0.5 to 84.4\%, resulting in a 4.2\% gain. This demonstrates that the EMA module effectively captures global information to learn richer semantic features, focusing the network model more on the overall context of retail product targets and enhancing model performance.

The final improved model, MidState-YOLO-ED, showed improvements in all evaluation metrics relative to YOLOv8-n, with precision and recall increasing by 2.3 and 1.6 percentage points, respectively, and \textit{mAP} reaching 89\%. Additionally, the number of parameters and floating point operations were significantly lower.

\subsection{Experimental Results and Discussion}

\begin{table*}[h]
  \centering
  \caption{Compare experimental data}
  \label{tab:table1}
  \begin{tabular}{@{}cccccc@{}}
  \hline
    Model            &	Recall         & mAP@0.5           & mAP@0.5:0.95        & Params            & GFLOPs \\
  \hline
    Faster R-CNN     &	0.899          &	\textbf{0.995} &	0.855            &	41808406         &	134.9 \\
    SSD              &	0.758          &	0.943          &	0.693            &	30160468         &	13.4 \\
    YOLOv8-n         &	0.981          &	0.992          &	0.869            &	3371024          &	9.8 \\
    YOLO v10-n       &	0.970          &	0.991          &	0.871            &	\textbf{2885888} &	\textbf{9.2} \\
    MidState-YOLO    &	\textbf{0.987} &	0.993          &	\textbf{0.875}   &	3405456          &	9.8 \\
    MidState-YOLO-ED & 	0.985          &	0.994          &	\textbf{0.875}   &	3288096          &	9.6 \\
  \hline
  \end{tabular}
\end{table*}
The experimental results primarily organize the performance parameters of the trained algorithms, explaining the strengths and weaknesses of each algorithm based on these results, and analyzing the experimental data and actual detection effects. All experiments were conducted under the same configuration settings, and after training, the weight files generated by each algorithm were tested. The algorithms used in the comparative experiments include SSD, Faster-RCNN, YOLOv8-n, and YOLOv10-n. The experimental results are shown in Table 3.

The experimental results indicate that, compared to the SSD and Faster-RCNN algorithms, the YOLO series algorithms and the improvements introduced in this study exhibit superior detection performance. Additionally, since Faster-RCNN is a two-stage algorithm, its applicability is limited by its complexity and the extended duration required for detection, which makes its overall performance inferior to that of the YOLO series.

The comparative results of different algorithms demonstrate that the MidState-YOLO-ED algorithm excels in terms of the number of parameters and floating point operations, with only 3,288,096 parameters and 9.6 GFLOPs. This fully proves the excellent performance of the improved algorithm in terms of lightweight design. The algorithm can process image data quickly and accurately, making it suitable for scenarios requiring fast response. It is also more appropriate for operation in resource-constrained environments such as mobile devices and embedded systems, and it meets real-time requirements\cite{ouyang2023efficient}.

Furthermore, key indicators such as recall and mAP for the MidState-YOLO-ED algorithm are higher than those of baseline algorithms, offering a new option for efficient and rapid object detection.

After training the final improved model, MidState-YOLO-ED, for 30 epochs, the prediction results are displayed in Figure 3. The results demonstrate that the algorithm proposed in this paper achieves excellent detection performance while maintaining a lightweight framework.
\begin{figure}
\centering
\includegraphics[width=0.5\textwidth]{"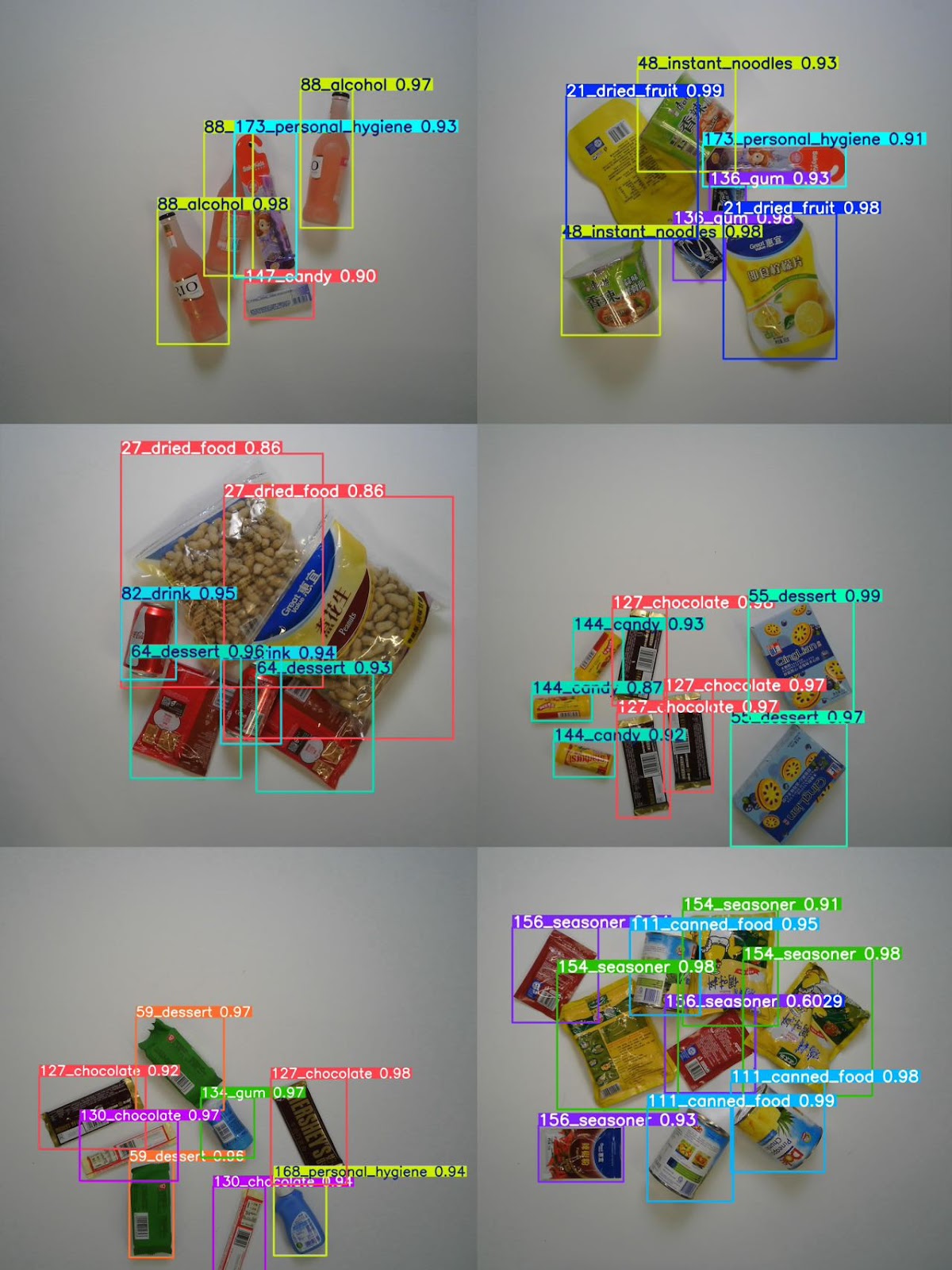"}
\caption{Dual Convolution C2f-Dual Design}
\end{figure}

\section{Conclusion}
In this paper, we presented an enhanced self-checkout system using an improved YOLOv10 network. The system significantly advances retail automation by optimizing checkout efficiency and minimizing labor costs. Our adaptations to the YOLOv10 model, integrating features from YOLOv8 and new post-processing algorithms, markedly improves product recognition accuracy, with our experiments demonstrating superior performance over existing systems. Broader applications in inventory control and customer service will be benefited by this study. Our study shows that AI-driven technologies will play a pivotal role in enhancing consumer experiences and operational efficiency.

\bibliographystyle{IEEEtran}
\bibliography{YOLO_paper/latex/YOLO_Manuscript}





\end{document}